\documentclass{sig-alt-release2}

\newcommand{\commentaire}[1]{ } 
\newcommand{\Sol}[0]{{\cal S}} 
\newcommand{\Vois}[0]{{\cal V}} 
\newcommand{\Real}{\mathop{\rm I\kern-.2emR}}
\renewcommand{\P}{\mathop{\rm I\kern-.2emP}}

\begin{document}

\conferenceinfo{GECCO'06,} {July 8--12, 2006, Seattle, Washington, USA.} 
\CopyrightYear{2006} 
\crdata{1-59593-186-4/06/0007} 

\newdef{definition}{Definition}

\title{Measuring the Evolvability Landscape to study Neutrality} 

%
%

\numberofauthors{3}

%

\author{
%
\alignauthor S. Verel\\
       \affaddr{Univ. of Nice-Sophia Antipolis}\\
       \affaddr{2000, route des Lucioles}\\
       \affaddr{Sophia, France}\\
       \email{verel@i3s.unice.fr}
\alignauthor P. Collard\\
       \affaddr{Univ. of Nice-Sophia Antipolis}\\
       \affaddr{2000, route des Lucioles}\\
       \affaddr{Sophia, France}\\
       \email{pc@i3s.unice.fr}
\alignauthor M. Clergue\\
       \affaddr{Univ. of Nice-Sophia Antipolis}\\
       \affaddr{2000, route des Lucioles}\\
       \affaddr{Sophia, France}\\
       \email{clerguem@i3s.unice.fr}
}

\date{}

\maketitle

\begin{abstract}

This theoretical work defines the measure of \textit{autocorrelation of evolvability}
in the context of neutral fitness landscape.
This measure has been studied on the classical MAX-SAT problem.
This work highlight a new characteristic of neutral fitness landscapes
which allows to design new adapted metaheuristic.

\end{abstract}



\section*{Introduction}

Fitness landscape, 
introduce by Wright \cite{stadler-02} in evolutionary biology,
is one of the powerful metaphor to model evolutionary process.
The dominant view in this metaphor is an adaptive evolution 
where an uphill walk of a population on a mountainous fitness landscape 
in which it can get stuck on suboptimal peaks.
In combinatorial optimization,
this view also influence the design of metaheuristic:
the geometry of multimodality or ruggedness describes the fitness landscape
and the metaheuristics try to escape from local optima
by using a probability to explore the landscape in simulated annealing,
by using a memory in tabu search,
or by preserving the diversity of population in evolutionary algorithm.

Another geometry of fitness landscapes, 
enlightened in molecular evolution by Kimura \cite{KIM:83}, 
takes an important place in optimization: the neutral fitness landscape.
In the theory of neutral evolution,
the overwhelming majority of mutations are 
either effectively neutral or lethal and in the latter case purged by 
negative selection.
According to this theory,
fitness landscapes are dominated by plateaus,
called \textit{neutral networks}.
The landscapes from genetic programming 
or from applicative problems
such as minimal linear arrangement
are known to be neutral.
It is difficult to decide whether neutrality is useful for optimization.
This work propose to deeper describe neutral fitness landscapes 
in order to get some characteristics of theses problems and
obtain a more complete view on neutrality 
which allows to design more useful metaheuristic.

\section{Fitness landscapes}
\label{subsec-fitnes-land}

We will use the definition of fitness landscapes from \cite{stadler-02}.

A {\it fitness landscape} is a triplet $(\Sol, \Vois, f)$ such as:
$\Sol$ is the set of potential solutions,
$\Vois : \Sol \rightarrow 2^\Sol$ is the neighborhood function 
which associated to each solution $s \in \Sol$ a set of neighbor solutions $\Vois(s) \subset S$,
and 
$f : \Sol \rightarrow \Real$ is the fitness function 
which associates a real number to each solution.

\subsection{Rugged Fitness landscapes}

Weinberger \cite{WEI:90} introduced the autocorrelation function and 
the correlation length of random walks to measure the ruggedness of fitness landscapes. \\
Given a random walk $( s_t, s_{t+1}, \ldots  )$,
the autocorrelation function $\rho$ of a fitness function $f$ is the autocorrelation
function of time series $( f(s_t), f(s_{t+1}), \ldots )$ :
$$
\rho(k) = \frac{E[f(s_t) f(s_{t+k})] - E[f(s_t)]E[f(s_{t+k})]}{var(f(s_t))}
$$
where $E[f(s_t)]$ and $var(f(s_t))$ are the expected value and the variance of $f(s_t)$.
The correlation length $\tau = - \frac{1}{log(\rho(1))}$,
measures how the autocorrelation function decreases 
and it summarizes the ruggedness of the landscape: 
the larger the correlation length, the smoother is the landscape.

\subsection{Neutral fitness landscapes}

The geometry of neutral fitness landscapes 
are based on the concept of neutral neighborhood and neutral networks.

For every $s \in S$, 
the \textit{neutral neighborhood} of $s$ is the set 
$\Vois_{neut}(s) = \lbrace s^{'} \in \Vois(s) ~|~ f(s) = f(s^{'}) \rbrace$ and 
the \textit{neutral degree} of $s$
is the number of neutral neighbors of $s$. 
There is no exact definition of neutral fitness landscape
but we can define a fitness landscape as neutral 
if there are ``many'' solutions with ``high'' neutral degree.
In this case, 
we can imagine fitness landscapes with some plateaus called \textit{neutral networks}. 
There is no significant difference of fitness between solutions on neutral networks 
and the population drifts around on them.

The Neutral Networks (NN) of a fitness landscape 
are connected graphs which are the connected components of graph $(\Sol, \Vois_{neut})$.
Solutions in a NN have the same fitness value 
and there is a path of neutral neighbors between two solutions of a NN.
The presence of NN takes place in the search dynamic.
For example, 
the time of drift on NN during the search process
depends on the properties of the networks.

\vfill\eject

\section{The autocorrelation of \\ evolvability}
\label{sec-autocor_evol}

\textit{Evolvability} is defined by Altenberg \cite{WA-AL} as "the ability of random variations to sometimes produce improvement". 
As enlighten by Turney~\cite{turney99increasing} the concept of evolvability is difficult to define. 
As he puts it: 
"if $s$ and $s{'}$ are equally fit, 
$s$ is more evolvable than $s{'}$ 
if the fittest offspring of $s$ is more likely to be fitter than 
the fittest offspring of $s{'}$". 
Following this idea the evolvability of a solution is defined by a function $ef$ that assigns to every $s \in {\cal S}$ a real number which measure the evolvability.
For examples, 
the evolvability function could be the maximum fitness from the neighborhood 
$ef(s) = max \lbrace f(s^{'}) ~|~ s^{'} \in  {\cal V}(s) \rbrace$.

We define the \textit{autocorrelation of evolvability} for a neutral network $N$ 
as the autocorrelation function of evolvability on neutral networks.

The {\it autocorrelation of evolvability} on the neutral network $N$
is the autocorrelation of series $( ef(s_0), ef(s_1), \ldots )$\\
where $(s_0, s_1, \ldots)$ is a neutral random walk on $N$.

A neutral random walk is a series of solutions $(s_0, s_1, \ldots)$
such as $s_{i+1} \in \Vois(s_i)$ and $f(s_{i+1}) = f(s_i)$.
To extend this measure to the set of all neutral networks,
the average of autocorrelation coefficient is computed.
Several choices could be made to define evolvability function;
in particular 
we call \textit{autocorrelation of maximal evolvability} 
the autocorrelation when the evolvability is 
the maximum fitness from the neighborhood of a solution.

The evolvability gives
the fitnesses of neutral network in the neighborhood.
For example, 
the maximum evolvability of a solution is the fitness of the higher NN in the neighborhood.
The autocorrelation of evolvability allows to describe the distribution of neutral networks around.
If the correlation is large,
the fitness in the neighborhood of a NN is distributed regularly,
whereas if the correlation is low,
the NNs around a NN is randomly distributed.

\section{First results on the MAX-k-SAT Problem}
\label{sec-maxsat}

In the following we present the first measures of autocorrelation of evolvability. 
The MAX-k-SAT is defined from the SAT optimization problem. 
SAT is a decision problem 
that asks whether a binary tuple can be found that satisfies all clauses in normal conjunctive form. 
Many studies deal with the solution space of the SAT problem \cite{Mezard02},
such phase transistion arround the threshold $\alpha_c$ defined as the ratio between the number of clauses and the number of variables.
For $k=3$, $\alpha_c$ is equal to $4.3$.

The experiments are led in the same way as for preview landscapes
with random instances of MAX-3-SAT. 
The number of variables is set to $N=16$ and $N=64$,
the number of literals by clause is $k=3$ and 
the number of clauses $m$ describe respectively 
the sets $\{39, 59, 64, 69, 74, 79, 99 \}$ 
and $\{ 200, 250, 265, 275, 285, 300, 350 \}$.
The average neutral degree decreases when $m$ increases.

\begin{figure}[!ht]
\begin{center}
\begin{tabular}{c}
  \epsfig{figure=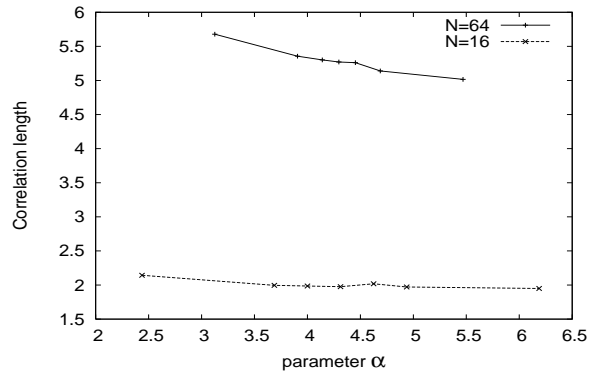   ,width=8cm,height=5cm} \\ 
\end{tabular}
\end{center}
\caption{Correlation length of maximal evolvability
for $N=16$ and $64$ and different number of clauses. 
\label{fig-maxsat-length} }
\end{figure}

The figure \ref{fig-maxsat-length} 
shows the correlation length of maximal evol-vability.
For all the parameters,
the correlation is significant.
The correlation length is around $2$ for $N=16$ 
and around $5$ for $N=64$.
For all the value of the number of clauses,
the autocorrelation functions are nearby
and the variations of correlation length 
are weak according to the parameter $\alpha$.
The correlation decreases slowly
according to the number of clauses,
which is linked to the neutrality.
The autocorrelation of maximal evolvability do not shows
a phase transition around the threshold $\alpha_c=4.3$.
The neutral networks are not randomly distributed
in the fitness landscapes.
Those first experiments show that the autocorrelation of evolvability
is one of characteristic of applicative neutral fitness landscapes.

\section{Conclusion}

From the metaphor of neutral fitness landscapes,
the neutral networks (NN) are the plateaus.
For each NN,
we have defined the autocorrelation function of evolvability.
The first studies on applicative MAX-SAT problem
has showed a new characteristic of those problem
which can be compared to the preview additive landscapes.

The autocorrelation of evolvability is 
an useful measure which highlight a new characteristic of neutral fitness landscapes
which could be study in real optimization problems.
In spite of a lack of differential between fitness in a NN,
evolvability could be exploit to guide the search process on a network.
Futures works could take into account this information to design new metaheuristics adapted to neutral fitness landscapes.

\bibliographystyle{abbrv}


\begin{thebibliography}{1}

\bibitem{WA-AL}
G.~Wagner.
\newblock Complexes adaptations and the evolution of evolvability.
\newblock In {\em Evolution}, pages 967--976, 1996.

\bibitem{KIM:83}
M.~Kimura.
\newblock {\em The Neutral Theory of Molecular Evolution}.
\newblock Cambridge University Press, Cambridge, UK, 1983.

\bibitem{Mezard02}
M.~Mezard and R.~Zecchina.
\newblock The random k-satisfiability problem: from an analytic solution to an
  efficient algorithm.
\newblock {\em Phys. Rev. E}, 66(056126), 2002.

\bibitem{stadler-02}
P.~F. Stadler.
\newblock Fitness landscapes.
\newblock In M.~L\"assig and Valleriani, editors, {\em Biological Evolution and
  Statistical Physics}, volume 585 of {\em Lecture Notes Physics}, pages
  187--207, Heidelberg, 2002. Springer-Verlag.

\bibitem{turney99increasing}
P.~D. Turney.
\newblock Increasing evolvability considered as a large scale trend in
  evolution.
\newblock In P.~Marrow and al, editors, {\em GECCO'99 , Workshop Program on
  evolvability}, pages 43--46, 1999.

\bibitem{WEI:90}
E.~D. Weinberger.
\newblock Correlated and uncorrelatated fitness landscapes and how to tell the
  difference.
\newblock In {\em Biological Cybernetics}, pages 63:325--336, 1990.

\end{thebibliography}

\end{document}